\documentclass[runningheads]{llncs}
\usepackage{tikz}
\usepackage{graphicx}
\usepackage{xcolor}
\usepackage{multirow, booktabs}
\usepackage{amssymb}
\usepackage{graphicx}
\usepackage{caption}
\usepackage{subcaption}
\usepackage{makecell}
\usepackage{caption}
\usepackage{tikz}
\usetikzlibrary{decorations.markings}
\usepackage{tabularx}
\usepackage{bm}

\usepackage[acronym]{glossaries}
\newacronym{ml}{ML}{Machine Learning}
\newacronym{wsi}{WSI}{Whole Slide Images}
\newacronym{aji}{AJI}{Aggregated Jaccard Index}
\newacronym{gan}{GAN}{Generative Adversarial Networks}

\usepackage{adjustbox}
\usetikzlibrary{shapes.geometric, arrows, positioning}

\tikzstyle{blue} = [rectangle, rounded corners, minimum width=8em, minimum height=1.8em,text centered, draw=black, fill=blue!30]
\tikzstyle{yellow} = [rectangle, rounded corners, minimum width=8em, minimum height=1.8em,text centered, draw=black, fill=yellow!30]
\tikzstyle{green} = [rectangle, rounded corners, minimum width=8em, minimum height=1.8em,text centered, draw=black, fill=green!30]
\tikzstyle{arrow} = [thick,->,>=stealth]

\usepackage{cleveref}

\begin{document}
\title{Realistic Data Enrichment for Robust Image Segmentation in Histopathology}
\titlerunning{Robust Image Segmentation in Kidney Transplant Pathology}

\author{***}
\institute{***}
\makeatletter
\def\@fnsymbol#1{\ensuremath{\ifcase#1\or \dagger\or \ddagger\or
   \mathsection\or \mathparagraph\or \|\or **\or \dagger\dagger
   \or \ddagger\ddagger \else\@ctrerr\fi}}
\newcommand*\samethanks[1][\value{footnote}]{\footnotemark[#1]}

\author{
Sarah Cechnicka \inst{1} \thanks{Equal contribution.}
\and
James Ball \inst{1} \samethanks
\and
Hadrien Reynaud \inst{1}
\and
Callum Arthurs \inst{2}
\and
Candice Roufosse \inst{2} 
\and
Bernhard Kainz \inst{1,3}
}

\authorrunning{S. Cechnicka et al.}
\institute{Departament of Computing, Imperial College London, UK 
\and
Centre for Inflammatory Disease, Imperial College London, London, UK
\and
Friedrich–Alexander University Erlangen–N\"urnberg, DE
\\
\email{sc7718@imperial.ac.uk}\\
}

\maketitle              % typeset the header of the contribution
\setcounter{footnote}{0} 
\begin{abstract}

Poor performance of quantitative analysis in histo\-pathological \gls{wsi} has been a significant obstacle in clinical practice.
Annotating large-scale \gls{wsi}s manually is a demanding and time-consuming task, unlikely to yield the expected results when used for fully supervised learning systems. Rarely observed disease patterns and large differences in object scales are difficult to model through conventional patient intake. Prior methods either fall back to direct disease classification, which only requires learning a few factors per image, or report on average image segmentation performance, which is highly biased towards majority observations. Geometric image augmentation is commonly used to improve robustness for average case predictions and to enrich limited datasets. So far no method provided sampling of a realistic posterior distribution to improve stability, \emph{e.g.} for the segmentation of imbalanced objects within images. Therefore, we propose a new approach, based on diffusion models, which can enrich an imbalanced dataset with plausible examples from underrepresented groups by conditioning on segmentation maps. Our method can simply expand limited clinical datasets making them suitable to train machine learning pipelines, and provides an interpretable and human-controllable way of generating histopathology images that are indistinguishable from real ones to human experts. We validate our findings on two datasets, one from the public domain and one from a Kidney Transplant study.\footnote{The source code and trained models will be publicly available at the time of the conference, on huggingface and github.}

%\keywords{segmentation  \and data augmentation \and histopathology \and image generation.}
\end{abstract}
\section{Introduction}
Large scale datasets with accurate annotations are key to the successful development and deployment of deep learning algorithms for computer vision tasks. Such datasets are rarely available in medical imaging due to privacy concerns and high cost of expert annotations. 
This is particularly true for histopathology,  where gigapixel images have to be processed \cite{synthetic_inRL}. This is one of the reasons why histopathology is, to date, a field in which image-based automated quantitative analysis methods are rare. 
In radiology, for example, most lesions can be characterised manually into clinically actionable information, \emph{e.g.} measuring the diameter of a tumour. However, this is not possible in histopathology, as quantitative assessment requires thousands of structures to be identified for each case, and most of the derived information is still highly dependent on the expertise of the pathologist. 
Therefore, supervised \gls{ml} methods quickly became a research focus in the field, leading to the emergence of prominent early methods~\cite{ronneberger2015u} and, more recently, to high-throughput analysis opportunities for the clinical practice~\cite{comppath,reisenbuchler2022local,han2017breast}. Feature location, shape, and size are crucial for diagnosis; this high volume of information required makes automatic segmentation essential for computational pathology~\cite{comppath}. 
%The extraction of these features is believed to lead digital pathology from human-led manual and empirical assessment of histopathology images to reproducible quantitative metrics-driven decision-making. 
The automated extraction of these features should lead to the transition from their time-consuming and error-prone manual assessment to reproducible quantitative metrics-driven analysis, enabling more robust decision-making.
% However, the frequent inability of \gls{ml} models to generalise across domains and to learn from small, imbalanced datasets~\cite{databalance} remains amongst the biggest challenges for digital pathology. 
% Therefore, in this work, we are interested in evaluating a training data generation approach in such a speciality domain --- on histopathology image analysis for kidney transplant biopsies. In order to maximize transplant survival and patient well-being, it is essential to identify and treat conditions that can result in graft failure, such as rejection, early on. The current diagnostic classification system presents difficulties for biopsy assessment due to its largely qualitative nature, high observer variability, and lack of granularity in crucial areas~\cite{van2021forecasting}. Histopathology evaluation of biopsies continues to be the gold standard for identifying organ transplant rejection~\cite{reeve2009diagnosing}. However, imbalances and small training sets still prevent deep learning methods from changing clinical practice in this field.
Evaluating biopsies with histopathology continues to be the gold standard for identifying organ transplant rejection~\cite{reeve2009diagnosing}. However, imbalances and small training sets still prevent deep learning methods from revolutionizing clinical practice in this field.

In this work, we are interested in the generation of training data for the specific case of histopathology image analysis for kidney transplant biopsies. In order to maximize transplant survival rates and patient well-being, it is essential to identify conditions that can result in graft failure, such as rejection, early on. The current diagnostic classification system presents shortcomings for biopsy assessment, due to its qualitative nature, high observer variability, and lack of granularity in crucial areas~\cite{van2021forecasting}.

\iffalse
%kers2022deep

%and expertise-dependant scoring with 

Traditional data augmentation techniques such as geometric and colour space transformations or image mixing attempt to mitigate data shortages. However, these often do not generalise well enough, especially with imbalanced datasets or cases out of distribution; thus, they may not be sufficient to achieve a balanced input \cite{databalance}. 

\fi

\noindent\textbf{Contribution:} We propose a novel data enrichment method using diffusion models conditioned on masks. Our model allows the generation of photo-realistic histopathology images with corresponding annotations to facilitate image segmentation in unbalance datasets or cases out of distribution. In contrast to conventional geometric image augmentation, we generate images that are indistinguishable from real samples to human experts and provide means to precisely control the generation process through segmentation maps. Our method can also be used for expert training, as it can  cover the extreme ends of pathological representations through manual definition of segmentation masks.

\noindent\textbf{Related Work:}
Diffusion Models have experienced fast-rising popularity%after their impressive results in text-to-image generation
~\cite{ramesh_zero-shot_2021,rombach_high-resolution_2022,saharia_photorealistic_2022}. Many improvements  have been proposed \cite{salimans_progressive_2022,song_denoising_2022}, some of them suggesting image-to-image transfer methods that can convert simple sketches into photo-realistic images~\cite{cheng2023adaptively}. This is partially related to our approach. However, in contrast to sketch-based synthesis of natural images, we aim at bootstrapping highly performing image segmentation methods from poorly labelled ground truth data. 

Data enrichment through synthetic images has been a long-standing idea in the community~\cite{databalance,InsMix,gupta2019gan}. So far, this approach was limited by the generative capabilities of auto-encoding~\cite{kingma2013auto} or generative adversarial approaches~\cite{goodfellow2020generative}. A domain gap between real and synthetic images often leads to shortcut learning~\cite{geirhos2020shortcut} and biased results with minimal gains. The best results have surprisingly been achieved, not with generative models, but with data imputation by mixing existing training samples to new feature combinations~\cite{dwibedi2017cut,tan2021detecting}. Sample mixing can be combined with generative models like \gls{gan} to enrich the data~\cite{InsMix}.  
%To the best of our knowledge, no existing methods are able to fully rely on synthetic data to train models that can outperform the state-of-the-art in automated segmentation of histopathology images. 

%This approach is most closely related to data augmentation techniques described by the latter three approaches. Due to larger images in its publically available dataset, the InsMix paper \cite{InsMix} and the methods it contrasted against will serve as a comparison. 

\section{Method}
\label{sec:meth}

We want to improve segmentation robustness. We denote the image space as $\mathcal{X}$ and label mask space as $\mathcal{Y}$. Formally, we look for different plausible variations within the joint space $\mathcal{X} \times \mathcal{Y}$ in order to generate extensive datasets $d_k = \{(\bm{x}_n^{(k)}, \bm{y}_n^{(k)})\}_{n=1}^{N_k}$, where ${N_k}$ is  the number of labelled data points in the $k$-th dataset. We hypothesise that training a segmentation network $M_\theta$ on combinations of $d_k$, 
$d_a \cup d_b \cup \dots \cup d_c$
with or without samples from an original dataset, will lead to state-of-the-art segmentation performance. We consider any image segmentation model $M_\theta: \mathcal{X} \rightarrow \mathcal{Y}$ that performs pixel-wise classification, \emph{i.e.} semantic segmentation, in $\mathbb{R}^{C}$, where $C$ is the number of classes in $\mathcal{Y}$. Thus, predictions for the individual segmentation class labels can be defined as $p(\bm{y}|\bm{x}, \theta) = \bm{\hat{y}} = softmax(M_\theta(\bm{x}))$.

Inverting the segmentation prediction to $p(\bm{x}|\bm{y}, \theta)$ is impractical, as the transformation $M_\theta$ is not bijective, and thus inverting it would yield a \emph{set} of plausible samples from $\mathcal{X}$.
However, the inversion can be modelled through a constrained sampling method, yielding single plausible predictions $\bm{\hat{x}} \in \hat{\mathcal{X}}$ given $\bm{y} \in \mathcal{Y}$ and additional random inputs $z \sim \mathcal{N}(0, \sigma) $ holding the random state of our generative process. 
Modelling this approach can be achieved through diffusion probabilistic models~\cite{DDPM}.%, described in \Cref{sec:diffmod}. 
We can thus define $D_{\phi}: \mathcal{Z} \rightarrow \hat{\mathcal{X}}$ where $\mathcal{Z}$ is a set of Gaussian noise samples. 
This model can be further conditioned on label masks $\bm{y}$ and produce matching elements to the joint space $\mathcal{X} \times \mathcal{Y}$ yielding $D_{\xi}:\mathcal{Z} \times \mathcal{Y} \rightarrow \hat{\mathcal{X}}$.

\begin{figure}
    \centering
    \includegraphics[width=\linewidth]{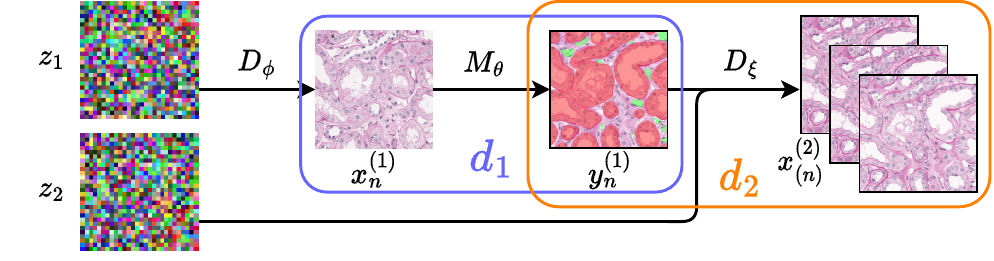}
    \caption{Summary of our dataset generation approach as described in Section \ref{sec:meth}. We use our diffusion model $D_{\phi}$ to generate images, $M_{\theta}$ segments them and $D_{\xi}$ creates multiple images from these segmentations. Dataset $d_2$ is the one used to train our final model $M_{\zeta}$.}
    \label{fig:method}
 %   \vspace{-0.5cm}
\end{figure}

The first step of our approach, shown in \Cref{fig:method}, is to generate a set of images $X_1=\{\bm{x}^{(1)}_n|\bm{x}^{(1)}_n=D_{\phi}(z), z \sim \mathcal{N}(0, \sigma)\} \subset \hat{\mathcal{X}}$ where $D_{\phi}$ is an unconditional diffusion model trained on real data samples. We then map all samples $\bm{x}^{(1)}_n$ to the corresponding elements in the set of predicted label masks $Y_1 = \{\bm{y}^{(1)}_n|\bm{y}^{(1)}_n=M_{\theta}(\bm{x}^{(1)}_n), \bm{x}^{(1)}_n~\in~X_1\} \subset \hat{\mathcal{Y}}$, where $M_{\theta}$ is a segmentation model trained on real data pairs. This creates a dataset noted $d_{1}$.
The second step is to generate a dataset $d_{2}$, by using a conditional diffusion model $D_{\xi}$ trained on real images and applied to the data pairs in $d_{1}$, such that $X_2=\{\bm{x}^{(2)}_n|\bm{x}^{(2)}_n=D_{\xi}(\bm{y}^{(1)}_n, z), \bm{y}^{(1)}_n\in Y_1, z \sim \mathcal{N}(0, \sigma)\}$. This lets us generate a much larger and more diverse dataset of image-label pairs, where the images are generated from the labels. 
Our final step is to use this dataset to train a new segmentation model $M_\zeta$ that largely outperforms $M_\theta$. To do so, we first train $M_\zeta$ on the generated dataset $d_{2}$ and fine-tune it on the real dataset.

%\subsection{Data Generation}

%We use two diffusion models to generate photorealistic pathological images.  ...

%-------------------------------------------
\noindent\textbf{Image Generation:} \label{sec:diffmod}
Diffusion models are a type of generative model producing image samples from Gaussian noise. The idea is to reverse a forward Markovian diffusion process, which gradually adds Gaussian noise to a real image $\bm{x}_0$ as a time sequence $\{\bm{x}_t\}_{t=1...T}$. The probability distribution $q$ for the 
forward sampling process at time $t$ can be written as a function of the original sample image 
\begin{equation}
q\left(\mathbf{x}_t \mid \mathbf{x}_0\right)=\mathcal{N}\left(\mathbf{x}_t ; \sqrt{\bar{\alpha}_t} \mathbf{x}_0,\left(1-\bar{\alpha}_t\right) \mathbf{I}\right), q\left(\mathbf{x}_t \mid \mathbf{x}_s\right)=\mathcal{N}\left(\mathbf{x}_t ;\left(\alpha_t / \alpha_s\right) \mathbf{x}_s, \sigma_{t \mid s}^2 \mathbf{I}\right),
\end{equation}
where $\bar{\alpha}_t=\sqrt{1 /\left(1+e^{-\lambda_t}\right)}$ and $\sigma_{t \mid s}^2=\sqrt{\left(1-e^{\lambda_t-\lambda_s}\right) \sigma_t^2}$ parameterise the variance of the noise schedule, whose logarithmic signal to noise ratio $\lambda_t=\log \left(\alpha_t^2 / \sigma_t^2\right)$ is set to decrease with t \cite{imagen,Sohl_Dickstein}.
The joint distribution $p_{\theta}$ describing the corresponding reverse process is 
\begin{equation}
    p_\theta\left(\mathbf{x}_{0: T}\right):=p\left(\mathbf{x}_T\right) \prod_{t=1}^T p_\theta\left(\mathbf{x}_{t-1} \mid \mathbf{x}_t\right), \quad p_\theta\left(\mathbf{x}_{t-1} \mid \mathbf{x}_t\right):=\mathcal{N}\left(\mathbf{x}_{t-1} ; \mathbf{\mu}_\theta\left(\mathbf{x}_t, t, \mathbf{c}\right),\sigma_t\right),
\end{equation}
where $\mu_{\theta}$ is the parameter to be estimated, $\sigma_t$ is given and $\mathbf{c}$ is an additional conditioning variable. Distribution $p$ depends on the entire dataset and is modelled by a neural network. 
% define e-parametrization
\cite{DDPM} have shown that learning the variational lower bound on the reverse process is equivalent to learning a model for the noise added to the image at each time step. By modelling $\mathbf{x}_t=\alpha_t \mathbf{x_0}+\sigma_t \mathbf{\epsilon}$ with $\mathbf{\epsilon} \sim \mathcal{N}(\mathbf{0}, \mathbf{I})$ we aim to estimate the noise $\mathbf{\epsilon_{\theta}}(x_t,\lambda_t,\mathbf{c})$ in order to minimise the loss function 
\begin{equation}
    \mathcal{L}=\mathbb{E}_{\mathbf{\epsilon}, \lambda_t,\mathbf{c}}\left[w(\lambda_t)\left\|\mathbf{\epsilon_{\theta}}(x_t,\lambda_t,\mathbf{c})-\mathbf{\epsilon}\right\|^2\right], 
\end{equation}
where $w(\lambda_t)$ denotes the weight assigned at each time step \cite{classifier_free_guidance}.
We follow~\cite{imagen} using a cosine schedule and DIMM~\cite{song_denoising_2022} continuous time steps for training and sampling.
%Classifier free guidance: 
We further use classifier free guidance~\cite{classifier_free_guidance} avoiding the use of a separate classifier network. The network partly trains using conditional input and partly using only the image such that the resulting noise is a weighted average:  
\begin{equation}
\tilde{\mathbf{\epsilon}}_\theta\left(x_t,\lambda_t,\mathbf{c}\right)=(1+w) \mathbf{\epsilon}_\theta\left(x_t,\lambda_t,\mathbf{c}\right)-w \mathbf{\epsilon}_\theta\left(\mathbf{z}_\lambda\right).
\end{equation}

The model can further be re-parameterized using 
 v-parameterization~\cite{salimans_progressive_2022}  by predicting $\mathbf{v}\equiv\alpha_t\epsilon-\sigma_t\mathbf{x}$ rather than just the noise, $\epsilon$, as before. With v-parameterization, the predicted image for time step $t$ is now $\mathbf{\hat{x}}=\alpha_t\mathbf{z}_t-\sigma_t\mathbf{\hat{v}_\theta}(\mathbf{z}_t)$.

\noindent\textbf{Mask conditioning:}
%Compared with the total number of histopathology patches in our dataset, we have a comparatively small number of labelled patches. 
Given our proprietary set of histopathology patches, only a small subset of these come with their corresponding segmentation labels.
Therefore, when conditioning on segmentation masks, we first train a set of unconditioned cascaded diffusion models using our unlabelled patches. This allows the model to be pre-trained on a much richer dataset, reducing the amount of labelled data needed to get high-quality segmentation-conditioned samples. Conditioning is achieved by concatenating the segmentation mask, which is empty in pre-training, with the noisy image as input into each diffusion model, at every reverse diffusion step.
After pre-training, we fine-tune the cascaded diffusion models on the labelled image patches so that the model learns to associate the labels with the structures it has already learnt to generate in pre-training.

\noindent\textbf{Mask Generation:}
We use a nnU-Net~\cite{nnU-Net} to generate label masks through multi-class segmentation. The model is trained through a combination of Dice loss $\mathcal{L}_{Dice}$ and Cross-Entropy loss $\mathcal{L}_{CE}$. % as defined below.
%Let $g$ denote the ground truth label mask and $s$ the predicted segmentation mask. $N$ denotes the number of pixels in a given image and $C$ corresponds to the number of classes in the ground truth \cite{10.1007/978-3-031-08999-2_36}.
\iffalse
Sørensen-Dice index is the most commonly used metric for evaluating segmentation accuracy \cite{YEUNG2022102026}, its corresponding loss is defined as:

% \begin{equation}\label{eq:dice_loss}
% \mathcal{L}_{Dice} = 1 - \frac{2\Sigma_{i=1}^{N}\Sigma_{c=1}^{C}g_i^cs_i^c}{\Sigma_{i=1}^{N}\Sigma_{c=1}^{C}g_i^{2c} +\Sigma_{i=1}^{N}\Sigma_{c=1}^{C}s_i^{2c} }
% \end{equation}

\begin{equation}\label{eq:dice_loss}
\mathcal{L}_{Dice} = 1 - \frac{2\Sigma_{i=1}^{N}\Sigma_{c=1}^{C}g_{i,c}s_{i,c}}{\Sigma_{i=1}^{N}\Sigma_{c=1}^{C}g_{i,c}^{2} +\Sigma_{i=1}^{N}\Sigma_{c=1}^{C}s_{i,c}^{2} },
\end{equation}

\noindent where $s_{i,c}$ is the binary prediction of pixel $i$ for class $c$, and $g_{i,c}$ is the binary corresponding ground truth.
% indicates if a class label \textit{c} is the accurate classification of pixel \textit{i}. 
\fi
$\mathcal{L}_{Dice}$ 
is used in combination with a Cross Entropy Loss $\mathcal{L}_{CE}$ to obtain more gradient information during training \cite{nnU-Net}, by giving it more mobility across the logits of the output vector.
\iffalse
\begin{equation}
\mathcal{L}_{CE} = -\frac{1}{N}\Sigma_{i=1}^{N}\Sigma_{c=1}^{C}g_{i,c}\log s_{i,c}
\end{equation}
\fi
%Inspired by nnU-Net~\cite{nnU-Net}, 
Additional auxiliary Dice losses are calculated at lower levels in the model. The total loss function for mask generation can therefore be described with 

\begin{equation}
\mathcal{L} = \mathcal{L}_{Dice} + \mathcal{L}_{CE} + \beta(\mathcal{L}_{Dice_{1/2}}+\mathcal{L}_{Dice_{1/4}}),
\end{equation}

\noindent where $\mathcal{L}_{Dice_{1/2}}$ and $\mathcal{L}_{Dice_{1/4}}$  denote the dice auxiliary losses calculated at a half, and a quarter of the final resolution, respectively.

We train two segmentation models $M_{\theta}$ and $M_{\zeta}$. First, for $M_{\theta}$, we train the nnU-Net on the original data and ground truth label masks. $M_{\theta}$ is then used to generate the label maps for all the images in $d_1$, the pool of images generated with our unconditional diffusion model $D_{\phi}$. The second nnU-Net, $M_{\zeta}$, is pre-trained on our dataset $d_2$ and we fine-tune it on the original data to produce our final segmentation model.
%The model is retrained as the last step in the pipeline supplemented by diffusion model generated patches and their corresponding masks. 

\section{Evaluation} %TODO
\noindent\textbf{Datasets and Preprocessing:}
We use two datasets for evaluation. The first one is the public KUMAR dataset~\cite{7872382}, which we chose to be able to compare with the state-of-the-art directly. KUMAR consists of 30 \gls{wsi} training images and 14 test images of $1000 \times 1000$ pixels with corresponding labels for tissue and cancer type (\verb!Breast!, \verb!Kidney!, \verb!Liver!, \verb!Prostate!, \verb!Bladder!,  \verb!Colon!, and \verb!Stomach!). During training, each raw image is cropped into a patch of $256 \times 256$ and then resized to $64 \times 64$ pixels. Due to the very limited amount of data available, we apply extensive data augmentation, including rotation, flipping, color shift, random cropping and elastic transformations. 
% The random shift is not used due to images being cropped at random. 
However the baseline methods~\cite{InsMix} only use 16 of the 30 images available for training. %We use the same reduced subset for a fair comparison.

The second dataset is a proprietary collection of Kidney Transplant Pathology \gls{wsi} slides with an average resolution of $30000 \times 30000$ per slide. These images were tiled into overlapping patches of $1024 \times 1024$ pixels. For this work, 1654 patches, classified as kidney cortex, were annotated (glomerulies, tubules, arteries and other vessels) by a consultant transplant pathologist with more than ten years of experience and an assistant with 5 years of experience. Among these, 68 patches, belonging to 6 separate \gls{wsi}, were selected for testing, while the rest were used for training. The dataset also includes tabular data of patient outcomes    %(\verb!Functioning!, \verb!25%!, \verb!50%!, \verb!Graft_Loss!, \verb!DWGL!, and \verb!DWFG!) 
and history of creatinine scores before and after the transplant. We resize the $1024 \times 1024$ patches down to $64 \times 64$ resolution and apply basic shifts, flips and rotations to augment the data before using it to train our first diffusion model. We apply the same transformations but with a higher re-scaling of $256 \times 256$ for the first super-resolution diffusion model. The images used to train the second and final super-resolution model are not resized but are still augmented the same. We set most of our training parameters similar to the suggested ones in~\cite{imagen}, but use the creatinine scores and patient outcomes as conditioning parameters for our diffusion models.

%KUMAR
%Kidney

%Our main experiment was carried out on a Kidney Transplant Pathology dataset that consists of 428 \gls{wsi} of post-transplant kidney biopsies. These 

\noindent\textbf{Implementation:}
We use a set of three cascaded diffusion models similar to~\cite{imagen}, starting with a base model that generates $64 \times 64$ images, and then two super-resolution models to upscale to resolutions $256 \times 256$ and $1024 \times 1024$. Conditioning augmentation is used in super-resolution networks to improve sample quality.   
%
% Apart from much of the same set-up as Imagen~\cite{imagen}, we additionally utilize v-parameterization~\cite{salimans_progressive_2022} in the 64×64 to 256×256 and 256×256 to 1024×1024 models. This re-parameterizes the model by predicting $\mathbf{v}\equiv\alpha_t\epsilon-\sigma_t\mathbf{x}$ rather than just the noise, $\epsilon$, as before. With v-parameterization, the predicted image for time step $t$ is now $\mathbf{\hat{x}}=\alpha_t\mathbf{z}_t-\sigma_t\mathbf{\hat{v}_\theta}(\mathbf{z}_t)$. We find that this different parameterization allows for fewer time steps when sampling whilst offering the same quality for the super-resolution models. This is particularly important for our 256×256 to 1024×1024 model which is much slower to sample from. We opt for the same noise-based parameterization used by Imagen~\cite{imagen} in the base model as sampling speeds are less of an issue due to the smaller size of the image being generated, and we saw a significant reduction in quality when using v-parameterization with only 256 time steps.
%
In contrast to \cite{imagen}, we use v-parametrization~\cite{salimans_progressive_2022} to train our super-resolution models ($64 \times 64 \rightarrow 256 \times 256$ and $256 \times 256 \rightarrow 1024 \times 1024$). These models are much more computationally demanding at each step of the reverse diffusion process, and it is thus crucial to reduce the number of steps during sampling to maintain the sampling time in a reasonable range. We find v-parametrization to allow for as few as 256 steps, instead of 1024 in the noise prediction setting, for the same image quality, while also converging faster. We keep the noise-prediction setting for our base diffusion model, as sampling speed is not an issue at the $64 \times 64$ scale, and changing to v-parametrization with 256 time steps generates images with poorer quality in this case.
We use PyTorch v1.6 and consolidated \cite{nnU-Net,imagen} into the same framework. Three Nvidia A5000 GPUs are used to train and evaluate our models. All diffusion models were trained with over 120,000 steps. The kidney study segmentation models were trained for 200 epochs and fine-tuned for 25, the KUMAR study used 800 epochs and was fine-tuned for 300. 
Training takes about 10 days and image generation takes 60 s per image.
Where real data was used for fine-tuning this was restricted to 30\% of the original dataset for kidney images.
Diffusion models were trained with a learning rate of $1\mathrm{e}{-4}$ and segmentation models were pre-trained with a learning rate of $1\mathrm{e}{-3}$ which dropped to $3 \mathrm{e}{-6}$ when no change was observed on the validation set in  15 epochs. Through  $D_{\phi}$,  $M_{\theta}$ and  $D_{\xi}$ the number of synthetic samples matched the number of real ones.  All models used Adam optimiser. See the supplemental material for further details about the exact training configurations.

\noindent\textbf{Setup:} 
We evaluate the performance of nnU-Net~\cite{nnU-Net} trained on the data enriched by our method. We trained over 5 different combinations of training sets, using the same test set for metrics comparison, and show the results in Table~\ref{tab:sota}. First, we train a base nnU-Net solely on real image data, (1), before fine-tuning it, independently, twice: once with a mixture of real and synthetic images as (2), and once exclusively with synthetic images as (3). The $4^{th}$ and $5^{th}$ models correspond to nnU-Nets retrained from scratch using exclusively synthetic images as (4), and one further fine-tuned on real images as (5) in Table~\ref{tab:sota}.

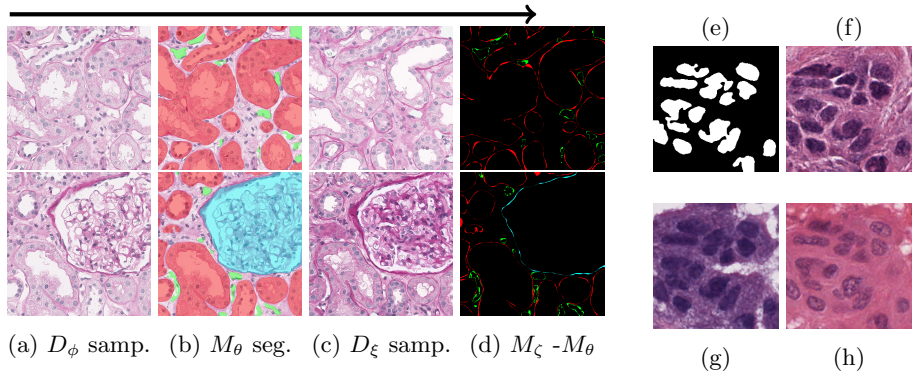
\begin{figure}[t]
\begin{minipage}{0.65\textwidth}
\begin{tikzpicture}
\coordinate (a) at (0,0);
\coordinate (b) at (7,0);
\draw (a) edge[->,line width=1.6pt] (b);
\end{tikzpicture}

\vspace{3pt}
        \begin{subfigure}[b]{0.24\textwidth}
         \centering
         \includegraphics[width=\textwidth]{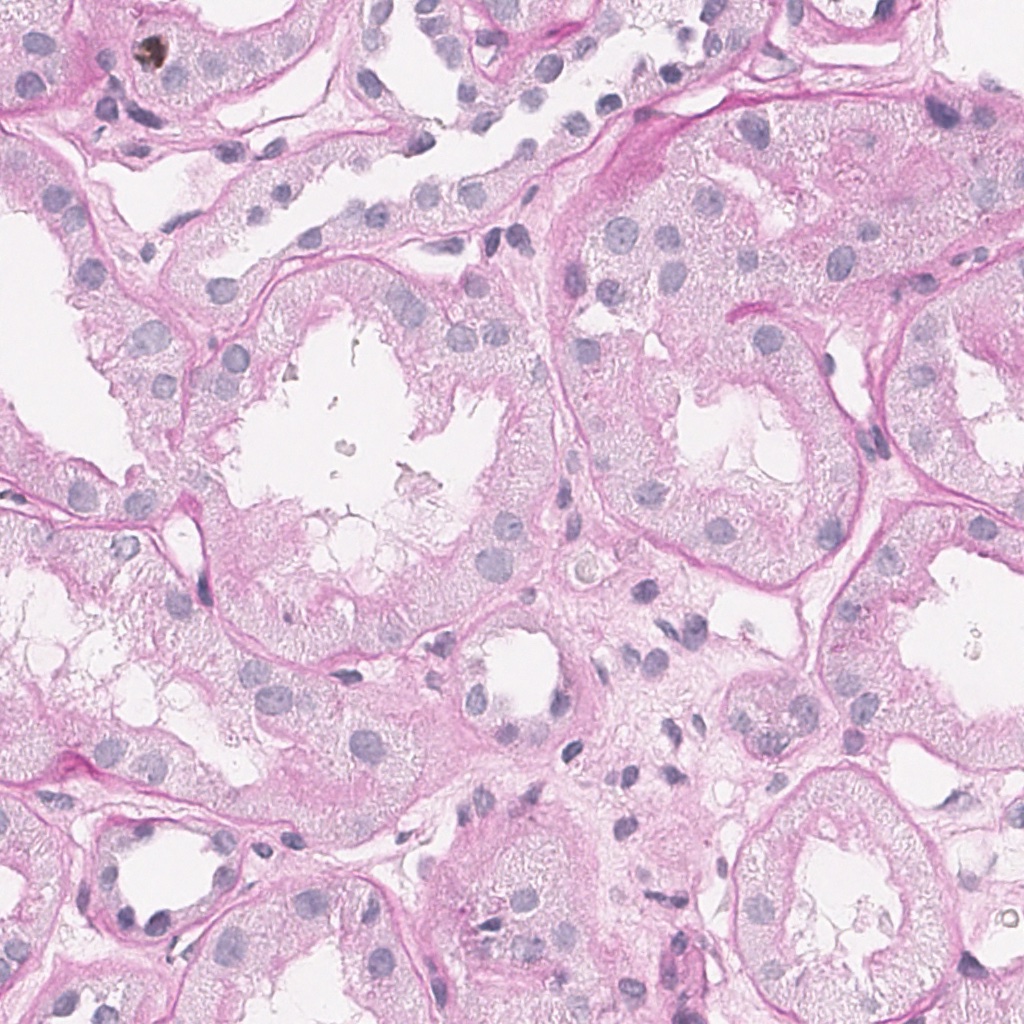}
         \includegraphics[width=\textwidth]{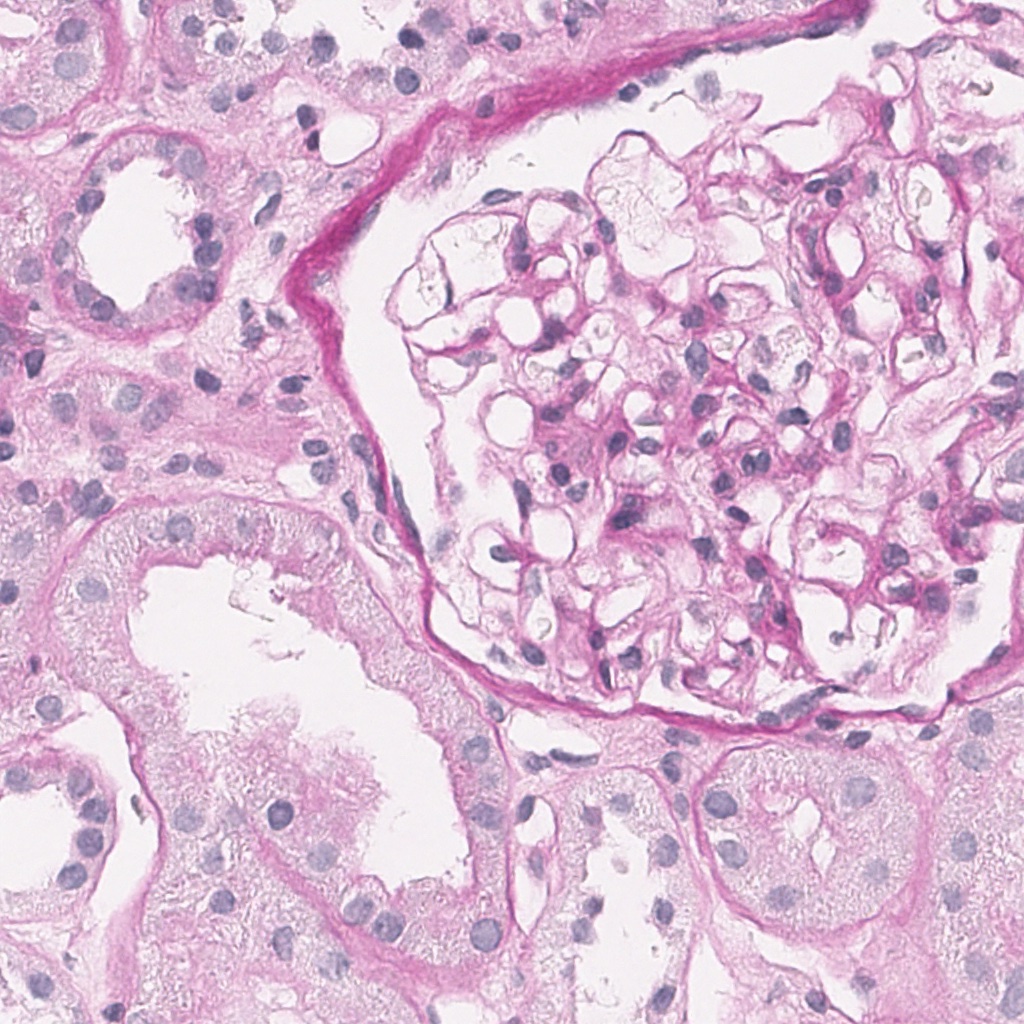}
                \caption{$D_{\phi}$ samp.}% \caption{orig.}
                \label{fig:mask}
        \end{subfigure}\hfill
        \begin{subfigure}[b]{0.24\textwidth}
                \centering
                \includegraphics[width=\textwidth]{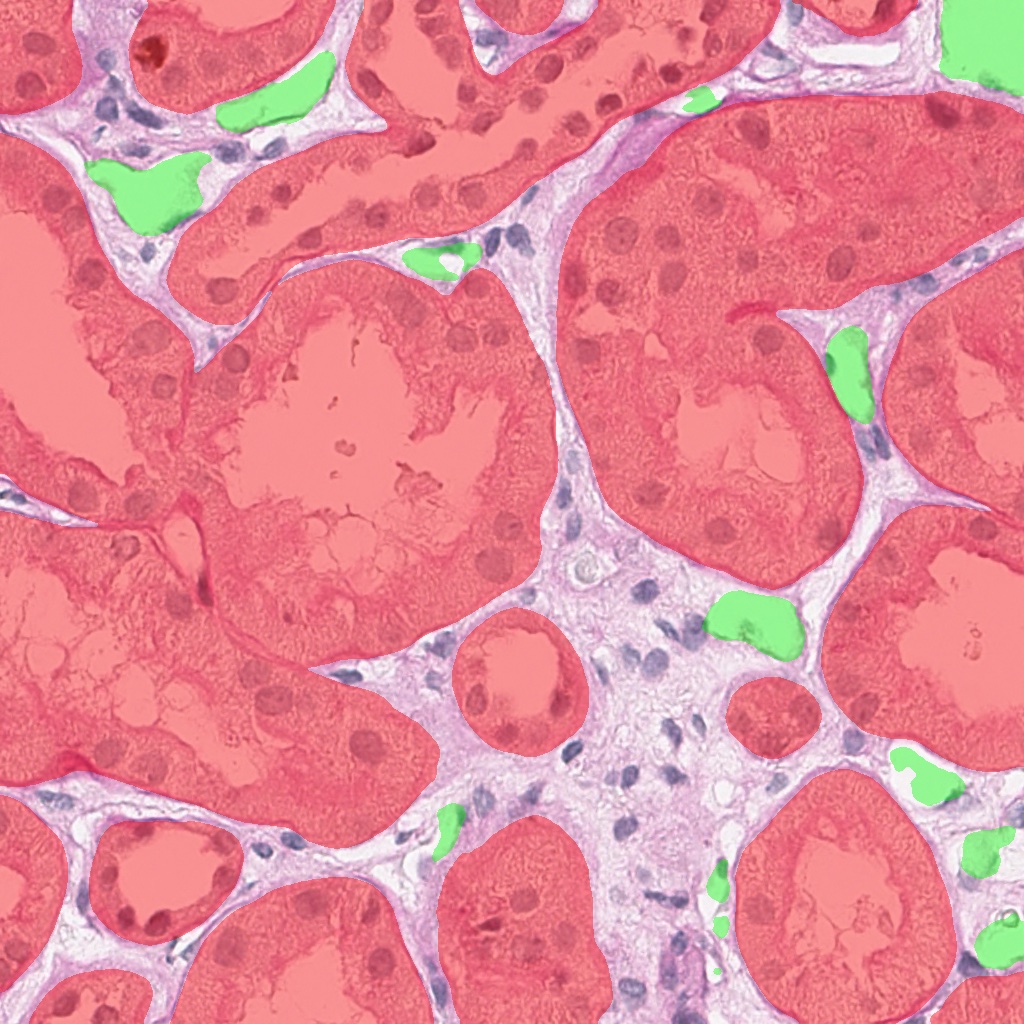}
                \includegraphics[width=\textwidth]{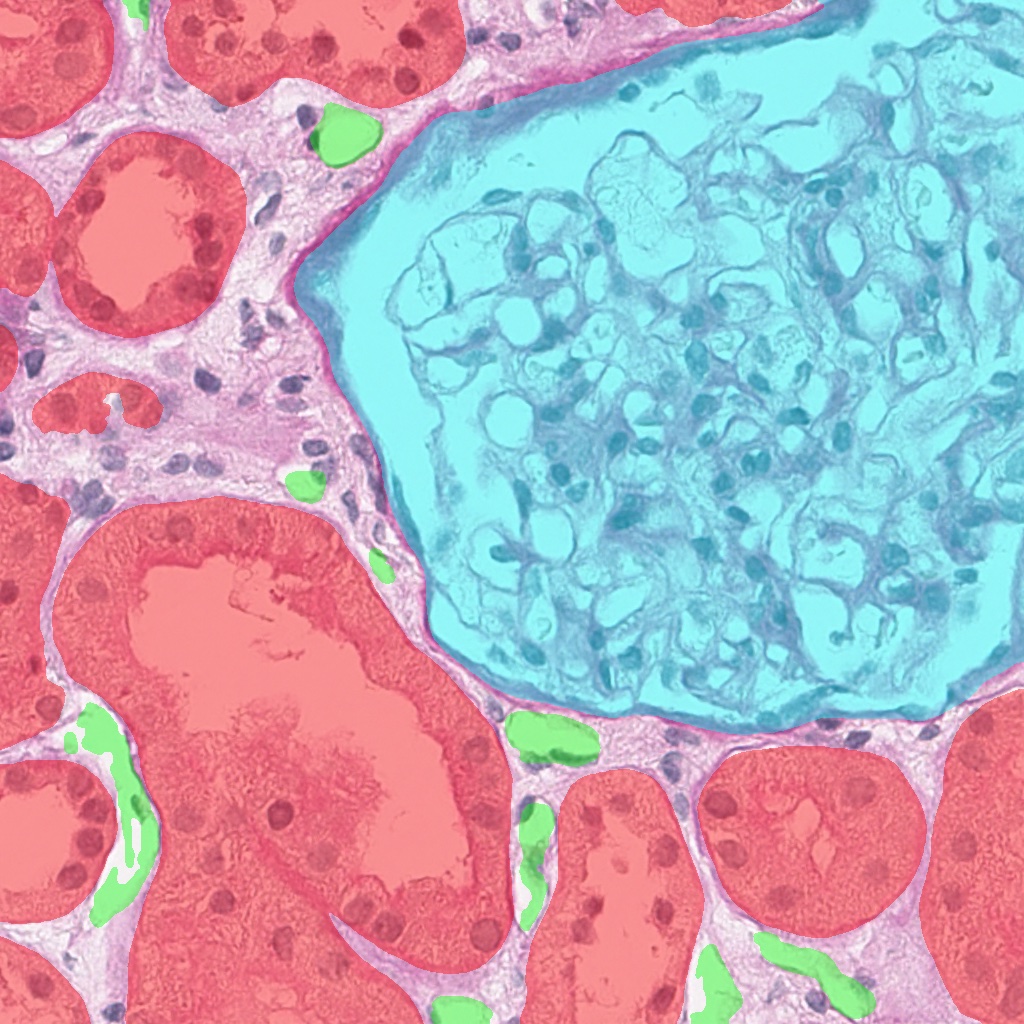}
                \caption{$M_{\theta}$ seg.}%\caption{labels}
                %Prostate adenocarcinoma
                \label{fig:X}
        \end{subfigure}\hfill
        \begin{subfigure}[b]{0.24\textwidth}
                \centering
                \includegraphics[width=\textwidth]{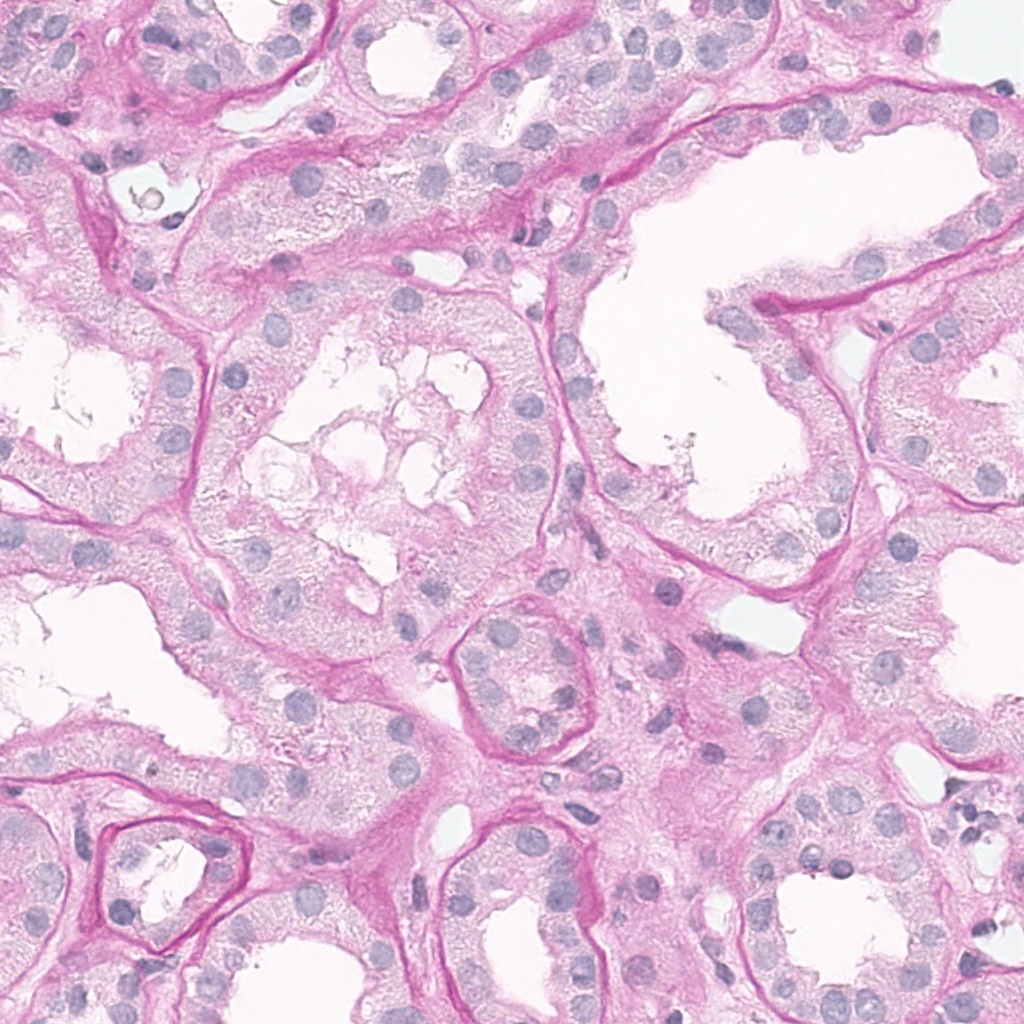}
                \includegraphics[width=\textwidth]{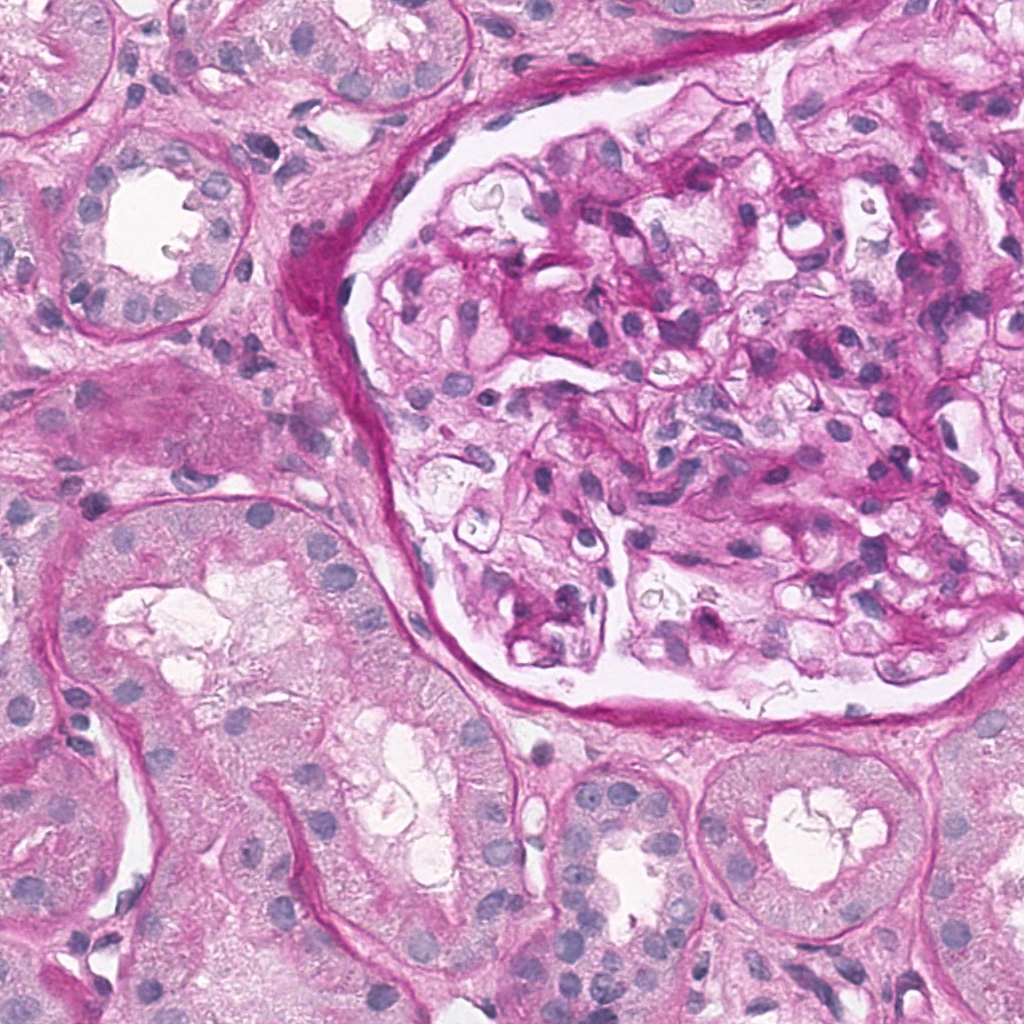}
                \caption{$D_{\xi}$ samp.}%\caption{gen.}
                %Stomach adenocarcinoma
                \label{fig:Y}
        \end{subfigure}\hfill
        \begin{subfigure}[b]{0.24\textwidth}
                \centering
                \includegraphics[width=\textwidth]{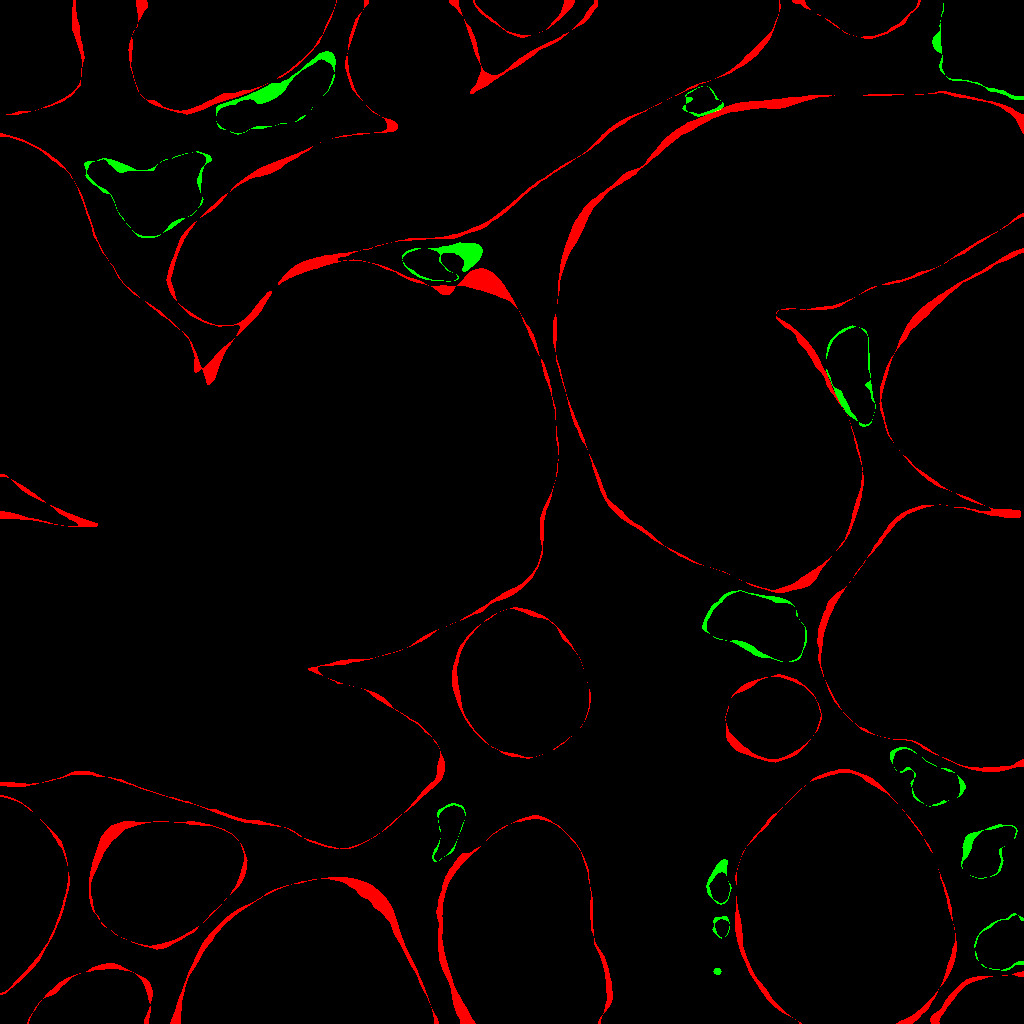}
                \includegraphics[width=\textwidth]{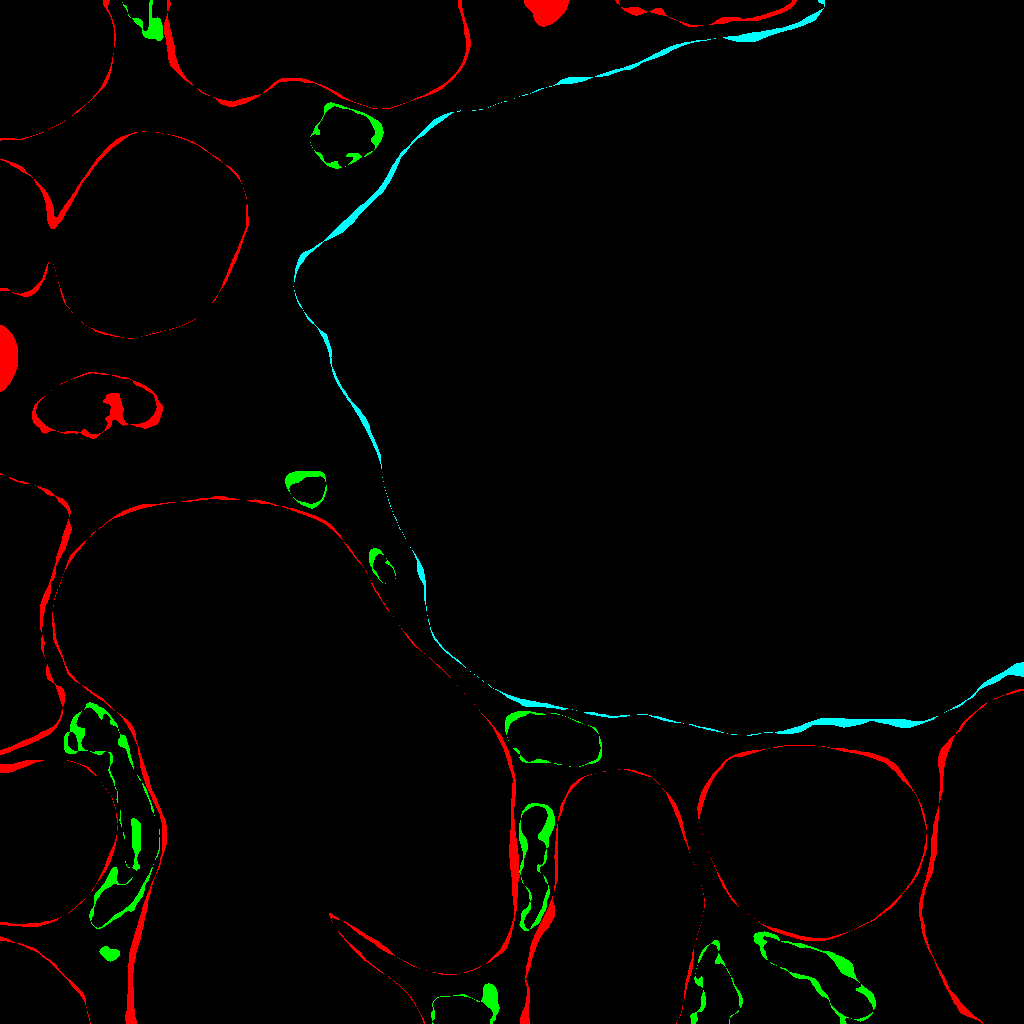}
                \caption{$M_{\zeta}$ -$M_{\theta}$}%\caption{diff.}
                %Stomach adenocarcinoma
                \label{fig:Z}
        \end{subfigure}%
        \end{minipage}~~~~~~~~
        \begin{minipage}{0.35\textwidth}
        \begin{subfigure}[b]{0.39\textwidth}
            \centering
            \caption{}
            \includegraphics[width=\textwidth]{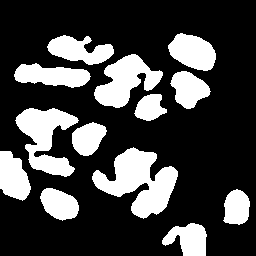}
            \label{fig:mask}
        \end{subfigure}
        \begin{subfigure}[b]{0.39\textwidth}
            \centering
            \caption{}
            \includegraphics[width=\textwidth]{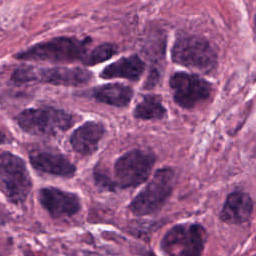}
            %Prostate adenocarcinoma
            \label{fig:X}
        \end{subfigure}
        \\
        \begin{subfigure}[b]{0.39\textwidth}
            \centering
            \includegraphics[width=\textwidth]{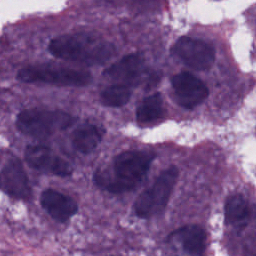}
            \caption{}
            %Stomach adenocarcinoma
            \label{fig:Y}
        \end{subfigure}
        \begin{subfigure}[b]{0.39\textwidth}
            \centering
            \includegraphics[width=\textwidth]{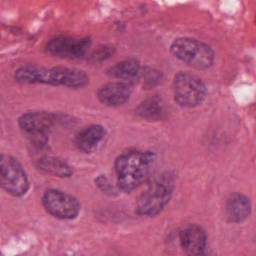}
            \caption{}
            %Lung adenocarcinoma
            \label{fig:Z}
        \end{subfigure}%
        \end{minipage}
        % \caption{Left: Examples from the data generation process. From left to right: original patch (a), segmentation label mask (b), generated image patch variant (c), and difference with label mask variant (d). red: tubule, blue: Glomeruli, green: vessels. Right: Diffusion model-generated images conditioned on different tissue types in KUMAR, using the same label mask (e). Generated images are for Liver (f), Bladder (g), and Breast (h). This example illustrates that our conditioning seems to allow any plausible label mask input from diverse sources.}
        \caption{Left: Outputs from our models. From left to right: (a) sample from $D_{\phi}$, (b) overlaid segmentation from $M_{\theta}$, (c) sample from $D_{\xi}$, (d) difference map of segmentation from $M_{\zeta}$ and $M_{\theta}$ highlighting shape improvement. Segmentation colors are: red: Tubuli, blue: Glomeruli, green: Vessels. Right: Diffusion model-generated images conditioned on different tissue types in KUMAR, using the same label mask (e). Generated images are from Liver (f), Bladder (g), and Breast (h) tissues. This shows that our conditioning seems to allow a plausible mask to produce any kind of tissue.}
        \label{fig:generatedimages}
%\vspace{-0.5cm}
\end{figure}

\iffalse
\begin{figure}[h]
        \begin{subfigure}[b]{0.24\textwidth}
         \centering
         \includegraphics[width=\textwidth]{images/43.png}
                \caption{Label  mask}
                \label{fig:mask}
        \end{subfigure}\hfill
        \begin{subfigure}[b]{0.24\textwidth}
                \centering
                \includegraphics[width=\textwidth]{images/12-gen-0.png}
                \caption{Liver}
                %Prostate adenocarcinoma
                \label{fig:X}
        \end{subfigure}\hfill
        \begin{subfigure}[b]{0.24\textwidth}
                \centering
                \includegraphics[width=\textwidth]{images/13-gen-1.png}
                \caption{Bladder}
                %Stomach adenocarcinoma
                \label{fig:Y}
        \end{subfigure}\hfill
        \begin{subfigure}[b]{0.24\textwidth}
                \centering
                \includegraphics[width=\textwidth]{images/14-gen-2.png}
                \caption{Breast}
                %Lung adenocarcinoma
                \label{fig:Z}
        \end{subfigure}%
        \caption{Diffusion model-generated images conditioned on different tissue types in KUMAR, but using the same label mask (a). This example illustrates that our conditioning seems to allow any plausible label mask input from diverse sources.}\label{fig:samemask}
\end{figure}
\fi
\noindent\textbf{Results:} 
Our quantitative results are summarised and compared to the state-of-the-art in Table~\ref{tab:sota} using the Dice coefficient (Dice) and \gls{aji} as suggested by~\cite{InsMix}. Qualitative examples are provided in Fig.~\ref{fig:generatedimages} (left), which illustrates that our model can convert a given label mask into a number of different tissue types and Fig.~\ref{fig:generatedimages}, where we compare synthetic enrichment images of various tissue types from our kidney transplant data.

\begin{table}[t]
\begin{center}
%will transpose that table if no other results from ours will be added
\caption{\label{tab:sota} Comparison with the state-of-the-art methods on the KUMAR dataset (top) and the limited KIDNEY transplant dataset (bottom). Metrics are chosen as in~\cite{InsMix}: Dice and \gls{aji}. Best values in bold.}
%\begin {adjustbox}{width=1.0\textwidth,center=\textwidth}
\begin{tabular}{c c l c c c p{0.1cm} c c c}
\toprule
    & & \multirow{2}{*}{Method} & \multicolumn{3}{c}{Dice (\%)}& &\multicolumn{3}{c}{AJI (\%)}\\ 
    & & & Seen & Unseen & All & & Seen & Unseen & All \\
    \cmidrule{4-6}\cmidrule{8-10}
    \multirow{12}{*}{\rotatebox{90}{KUMAR}} & & CNN3 \cite{CNN3Tab} & 82.26 & 83.22 & 82.67 & & 51.54 & 49.89 & 50.83\\  
    & & DIST \cite{DISTTab}  & - & - & - & &55.91 & 56.01 & 55.95\\  
    & & NB-Net \cite{NBNETTAB}  & 79.88 & 80.24 &  80.03 && 59.25 & 53.68 & 56.86\\  
    & & Mask R-CNN \cite{He_2017_ICCV}  & 81.07 & 82.91 & 81.86 & &59.78 &  55.31 & 57.86 \\  
    & & HoVer-Net \cite{HoVer} (*Res50)   & 80.60 & 80.41 & 80.52 && 59.35 &  56.27 & 58.03 \\  
    & & TAFE \cite{TAFE} (*Dense121)  & 80.81  &  83.72 & 82.06 && 61.51 & 61.54 & 61.52 \\  
    & & HoVer-Net + InsMix \cite{InsMix}  & 80.33 & 81.93 & 81.02 && 59.40 &  57.67 & 58.66 \\  
    & & TAFE + InsMix \cite{InsMix}   & 81.18 & 84.40 & 82.56 & & \textbf{61.98} &  \textbf{65.07} & \textbf{63.31} \\
    \cmidrule{3-10}   
     & \multirow{5}{*}{\rotatebox{90}{Ours}} & (1) trained on real & 82.97&84.89&83.52&&      52.34 &54.29& 52.90\\
    & & (2) fine-tuned by synthetic+real  & \textbf{87.82}& 88.66 & \textbf{88.06} && 60.79 & 60.05 & 60.71 \\  
    & & (3) fine-tuned by synthetic  & 87.12   &87.52 & 87.24 && 59.53 & 58.85 & 59.33\\  
    & & (4) trained on synthetic &86.06 &\textbf{89.69}  &87.10&& 52.89 &58.93 & 54.62 \\ 
    &&(5) \multirow{2}{*}{\makecell[l]{trained on synthetic,\\ fine-tuned on real}} & 85.75&  87.88&86.36&&56.01&57.83&56.5 \\ 
    &&&&&&&&&\\
    \midrule  
    %\midrule  
    \multirow{7}{*}{\rotatebox{90}{KIDNEY}}&\multirow{7}{*}{\rotatebox{90}{Ours}}%&(1) trained on real (100\% data) &  &  & 94.22 &&  &  & 77.45 \\   
    %&&(2) trained on real (30\% data)   & - & - & 75 && - & - & 40 \\ 
    %&&(2) trained on real (15\% data)   & - & - & 65 && - & - & 30 \\  &
    &(1) trained on real (30\% data) &  &  & 88.01 &&  &  & 62.05\\
   % &&(1b) trained on real (12\% data) & - & - & 80.75 && - & - & 50.23\\
    &&(2) fine-tuned by synthetic+real   &  &  & 92.25 &&  &  & 69.11 \\   
    &&(3) fine-tuned by synthetic   &  &  & 89.65 & &  &  & 58.59 \\  
    &&(4) trained on synthetic &  &  & 82.00 &&  &  & 42.40 \\ 
    &&(5) \multirow{2}{*}{\makecell[l]{trained on synthetic,\\ fine-tuned on real}} &  &  & \textbf{92.74} && & & \textbf{71.55} \\ 
    &&&&&&&&\\
    %\iffalse
  %  \cmidrule{3-10}  
    %&%\multicolumn{2}{c}{\multirow{2}{*}{\rotatebox{90}{\makecell[l]{\scriptsize{8.5\%} \\ \scriptsize{data}}}}}
    %\multirow{3}{*}{\rotatebox{90}{\makecell[c]{500 patches only}}}
   % &(6) trained on real (12\% data) & - & - & 80.75 && - & - & 50.23\\ 
    %&&(7) fine-tuned by synthetic+real& -& - & - && - & - & - \\ 
    %&&(8) \multirow{2}{*}{\makecell[l]{trained on synthetic,\\ fine-tuned on real}} & \textbf{-} & \textbf{-} & \textbf{} && \textbf{-} & \textbf{-} & \textbf{} \\ 
    %&&&&&&&&\\
    %\fi
\bottomrule
\end{tabular}
%\vspace{-0.5cm}
%\end{adjustbox}
\end{center}
\end{table}

\noindent\textbf{Sensitivity analysis:} 
Out of our 5 models relying on additional synthetic data in the KUMAR dataset experiments, all outperform previous SOTA on the Dice score. Importantly, synthetic results allow for high performance in previously unseen tissue types. 
Results are more nuanced when it comes to the AJI, as AJI over-penalizes overlapping regions~\cite{HoVer}. Additionally, while a further \gls{aji} loss was introduced to the final network $M_{\zeta}$, loss reduction, early stopping and the $M_{\theta}$ networks do not take it into account. 
Furthermore, Table~\ref{tab:sota} shows that, for the KIDNEY dataset, we can reach high performance (88\% Dice) while training $M_{\zeta}$ on $30\%$ (500 samples) of the real KIDNEY data (1). 
We also observe that the model pretrained on synthetic data and fine-tuned on 500 real images (5), outperforms the one only trained on 500 real images (1). 
Additionally, we discover that training the model on real data before fine-tuning it on synthetic samples (3) does not work as well as the opposite approach. 
We argue that pre-training an \gls{ml} model on generated data gives it a strong prior on large dataset distributions and alleviates the need for many real samples in order to learn the final, exact, decision boundaries, making the learning procedure more data efficient.

%As Table~ ref{tab:sota} also shows, the KIDNEY data saturates the model to very high performance, reaching $94.22\%$.  We test our model using only $12.5\%$ of the dataset, $192$ patches. We then fine-tuned using Model (2) from Table~\ref{tab:sota}, on all the synthetic images in addition to the patches already used for baseline training. We show that the model can reach a performance increase in combination with synthetic data. 

\noindent\textbf{Discussion:} We have shown that data enrichment with generative diffusion models can help to boost performance in low data regimes, \emph{e.g.}, KUMAR data, but also observe that when using a larger dataset, where maximum performance might have already been reached, the domain gap may become prevalent and no further improvement can be observed, \emph{e.g.}, full KIDNEY data (94\% Dice). Estimating the upper bound for the required labelled ground truth data for image segmentation is difficult in general. However, testing model performance saturation with synthetic data enrichment might be an experimental way forward to test for convergence bounds. Finally, the best method for data enrichment seems to depend on the quality of synthetic images. %Finally, we observe that pre-training on synthetic images and then fine-tuning on real images leads to the best performance in very data-limited scenarios, compared with other training strategies.

\section{Conclusion}
In this paper, we propose and evaluate a new data enrichment and image augmentation scheme based on diffusion models. We generate new, synthetic, high-fidelity images from noise, conditioned on arbitrary segmentation masks. This allows us to synthesise an infinite amount of plausible variations for any given feature arrangement. We have shown that using such enrichment can have a drastic effect on the performance of segmentation models trained from small datasets used for histopathology image analysis, thus providing a mitigation strategy for expensive, expert-driven, manual labelling commitments. \\

\noindent\textbf{Acknowledgements:} This work was supported by the UKRI Centre for Doctoral Training in Artificial Intelligence for Healthcare (EP/S023283/1). Dr. Roufosse is supported by the National Institute for Health Research (NIHR) Biomedical Research Centre based at Imperial College Healthcare NHS Trust and Imperial College London. The views expressed are those of the authors and not necessarily those of the NHS, the NIHR or the Department of Health. Dr Roufosse’s research activity is made possible with generous support from Sidharth and Indira Burman. The authors gratefully acknowledge the scientific support and HPC resources provided by the Erlangen National High Performance Computing Center (NHR@FAU) of the Friedrich-Alexander-Universität Erlangen-Nürnberg (FAU) under the NHR projects b143dc and b180dc. NHR funding is provided by federal and Bavarian state authorities. NHR@FAU hardware is partially funded by the German Research Foundation (DFG) – 440719683. Additional support was also received by the ERC - project MIA-NORMAL 101083647,  DFG KA 5801/2-1, INST 90/1351-1 and by the state of Bavarian.

\iffalse
\noindent\textbf{Acknowledgements:} 
This work is supported by the UKRI
Centre for Doctoral Training in Artificial Intelligence for Healthcare (EP/S023283/1). 
Dr. Roufosse is supported by the National Institute for Health Research (NIHR) Biomedical Research Centre based at Imperial College Healthcare NHS Trust and Imperial College London. The views expressed are those of the authors and not necessarily those of the NHS, the NIHR or the Department of Health. Dr Roufosse’s research activity is made possible with generous support from Sidharth and Indira Burman. The authors gratefully acknowledge the scientific support and HPC resources provided by the Erlangen National High Performance Computing Center (NHR@FAU) of the Friedrich-Alexander-Universität Erlangen-Nürnberg (FAU) under the NHR project  b143dc22. NHR funding is provided by federal and Bavarian state authorities. NHR@FAU hardware is partially funded by the German Research Foundation (DFG) 440719683. 
\fi
\newpage
\bibliographystyle{splncs04}
\bibliography{references}

\end{document}